\pdfoutput=1

\documentclass[11pt]{article}

\usepackage{ACL2023}

\usepackage{times}
\usepackage{latexsym}

\usepackage[T1]{fontenc}

\usepackage[utf8]{inputenc}

\usepackage{microtype}

\usepackage{inconsolata}

\usepackage{covington}

\usepackage{tabularx}
\usepackage{makecell}
\usepackage{multirow}
\usepackage{graphicx}
\usepackage{stfloats}

\title{\textit{Rarely} a problem? Language models exhibit inverse scaling in their predictions following \textit{few}-type quantifiers}

\author{James A. Michaelov \\
  Department of Cognitive Science \\
  University of California, San Diego \\
  \texttt{j1michae@ucsd.edu} \And
  Benjamin K. Bergen \\
  Department of Cognitive Science \\
  University of California, San Diego \\
  \texttt{bkbergen@ucsd.edu} }

\begin{document}
\maketitle
\begin{abstract}
How well do language models deal with quantification? In this study, we focus on \textit{few}-type quantifiers, as in \textit{few children like toys}, which might pose a particular challenge for language models because the sentence components without the quantifier are likely to co-occur, and \textit{few}-type quantifiers are rare. We present 960 English sentence stimuli from two human neurolinguistic experiments to 22 autoregressive transformer models of differing sizes. Not only do all the models perform poorly on \textit{few}-type quantifiers, but overall the larger the model, the worse its performance. This inverse scaling is consistent with previous work suggesting that larger models increasingly reflect online rather than offline human processing, and we argue that the decreasing performance of larger models may challenge uses of language models as the basis for natural language systems.  
\end{abstract}

\section{Introduction}
Quantifiers can dramatically alter the meaning of an utterance. Consider the sentences in (\ref{ex:sharks}).

\begin{subexamples}
\label{ex:sharks}
\item Most sharks are harmless.
\item Most sharks are dangerous.
\item Few sharks are harmless.
\item Few sharks are dangerous.
\end{subexamples}

Despite the fact that (a) and (c) have the same content words in the same syntactic arrangement, the statements have starkly different meanings. The same is true of (b) and (d). Being able to successfully comprehend these differences is useful, and in an example such as this one, vitally important\footnote{Note that most sharks are in fact harmless to humans; see, e.g., \url{https://www.floridamuseum.ufl.edu/discover-fish/sharks/shark-attack-faq/}.}.

Yet current work suggests that language models deal poorly with quantifiers---they struggle to predict which quantifier is used in a given context \citep{pezzelle_2018_ThemCanBe,talmor_2020_OLMpicsOnWhatLanguage}, and also perform poorly at generating appropriate continuations following logical quantifiers \citep{kalouli_2022_NegationCoordinationQuantifiers}. This is especially concerning given the recent trend of using large language models (sometimes referred to as `foundation models'; \citealp{bommasani_2021_OpportunitiesRisksFoundation}) as general systems that can perform multiple tasks, including question answering, without specific training \citep{brown_2020_LanguageModelsAre,raffel_2020_ExploringLimitsTransfer,lin_2021_FewshotLearningMultilingual,srivastava_2022_ImitationGameQuantifying,hoffmann_2022_TrainingComputeOptimalLarge,rae_2022_ScalingLanguageModels,zhang_2022_OPTOpenPretrained,chowdhery_2022_PaLMScalingLanguage}. It is thus crucial that such systems be able to distinguish among sentences like those in (\ref{ex:sharks}) in human-like ways both during training and when generating responses.

The aim of the present study is to evaluate how well language models take into account the meaning of a quantifier when generating the text that follows it, and to investigate whether this scales with model size. We are particularly interested in the question of whether language models exhibit \textit{inverse scaling}---that is, whether as model size increases, performance decreases rather than increases \citep{perez_2022_AnnouncingInverseScaling,mckenzie2022inverse}. Inverse scaling is an issue of serious concern for developing and training new language models, since inverse scaling could indicate `outer misalignment' \citep{perez_2022_AnnouncingInverseScaling}---that the training approach is leading to models that produce undesirable outputs, which may get worse as performance at training objectives increases. Inverse scaling is also a concern for models' ultimate use. As models increase in size and perform better at a wider range of benchmarks \citep[for recent examples, see, e.g.,][]{srivastava_2022_ImitationGameQuantifying,chowdhery_2022_PaLMScalingLanguage}, they may be increasingly assumed to be trustworthy and general-purpose, and thus able to perform well tasks on which they have not been tested \citep{raji_2021_AIEverythingWhole}. This could lead to a range of possible harms, from misidentifying whether something is dangerous or not (as in the opening example), to amplifying biases \citep{bender_2021_DangersStochasticParrots}.

To test how well language models deal with quantifiers, we follow the approach of \citet{ettinger_2020_WhatBERTNot} in using sentences from a study on human language comprehension to inform our evaluation. \citet{ettinger_2020_WhatBERTNot} found that following a negation, the predictions of BERT\textsubscript{\textsc{base}} and BERT\textsubscript{\textsc{large}} in simple sentences expressing a proposition with or without negation (from \citealp{fischler_1984_BrainPotentialsSentence}) do not appear sensitive to negation---for example, BERT\textsubscript{\textsc{large}} predicts the final word of \textit{a robin is a \textbf{bird}} to be more likely than \textit{a robin is a \textbf{tree}}, but also predicts that \textit{a robin is not a \textbf{bird}} is more likely than \textit{a robin is not a \textbf{tree}}. In this way, the models' predictions more closely match those made by humans `online'---that is, incrementally during the process of language comprehension---than our fully-formed `offline' judgements: in their original study, \citet{fischler_1984_BrainPotentialsSentence} found that the word \textit{bird} elicited an N400 response of smaller amplitude than \textit{tree} in both contexts, indicating that it was more strongly predicted.

Similar effects have been reported \citep{kassner_2020_NegatedMisprimedProbes,kalouli_2022_NegationCoordinationQuantifiers} for other transformers such as Transformer-XL \citep{dai_2019_TransformerXLAttentiveLanguage}, RoBERTa \citep{liu_2019_RoBERTaRobustlyOptimized}, and ALBERT \citep{lan_2020_ALBERTLiteBERT}, as well as ELMo \citep{peters_2018_DeepContextualizedWord}. Worse, recent work suggests that as language models increase in size, their ability to deal with negation may degrade: an inverse scaling relationship has been reported for performance at a wide range of tasks when prompts include negation \citep{mckenzie2022round1,jang_2023_CanLargeLanguage}, though it is possible that this may reverse at extremely large scales \citep{wei_2022_InverseScalingCan}.

Negation may be particularly challenging for statistical language models because its presence radically alters the meaning of a sentence, but negation occurs in only about 10\% of sentences \citep{jimenez-zafra_2020_CorporaAnnotatedNegation}. Quantifiers similarly impose radical modulations to meaning while also being relatively infrequent (see \autoref{sec:appendix_quantifiers}). In the present study, we focus on quantifiers indicating typicality such as \textit{most} and \textit{few}. To the best of our knowledge, only one study has evaluated model predictions following any quantifiers \citep{kalouli_2022_NegationCoordinationQuantifiers}, and it focused on words corresponding to logical quantifiers such as \textit{all}, \textit{every}, and \textit{some}. The few studies involving the quantifiers we address either focus on predicting the quantifier itself \citep{pezzelle_2018_ThemCanBe,talmor_2020_OLMpicsOnWhatLanguage}, or use RNNs to investigate modeling significant effects on the N400 without any form of evaluation \citep{michaelov_2020_HowWellDoes}. This study, therefore, represents the first attempt to explicitly evaluate the predictions of language models following \textit{most} and \textit{few}-type quantifiers.

In the present study, we carry out two experiments. In the first, following \citet{ettinger_2020_WhatBERTNot}, we use the stimuli from a previously published N400 study \citep{urbach_2010_QuantifiersMoreLess}. In it, \citet{urbach_2010_QuantifiersMoreLess} found that while \textit{most} and \textit{few}-type quantifiers do impact N400 amplitude, it is not enough to reverse predictions---\textit{few farmers grow \textbf{crops}} elicits a smaller N400 response than \textit{few farmers grow \textbf{worms}}, indicating that \textit{crops} was more strongly predicted than \textit{worms}, even though experimental participants judged it to be less plausible off-line. We test whether language models show the same pattern of insensitivity towards the quantifiers that humans do in online measures. In this way, we test how closely the predictions of language models correlate with those underlying the human N400 response. 

In our second experiment, we extend our study further. Experiment 1 aims to replicate the original N400 results of \citet{urbach_2010_QuantifiersMoreLess}; however, one thing that it does not account for is that while a given complete sentence (e.g., \textit{few farmers grow \textbf{crops.}}) can be highly unlikely and implausible, sentences beginning with the same words may not be (for example, in the plausible sentence \textit{few farmers grow \textbf{crops} in the winter}). Experiment 1 does not distinguish between these possibilities, and while it is important to test the sensitivity of language models to \textit{few}-type quantifiers, if they fail to show a difference for complete sentences including the final period (e.g., \textit{few farmers grow \textbf{crops.}}), this is more concerning. Thus, in Experiment 2, we run the same stimuli as Experiment 1, but including a period following the final word (e.g., \textit{crops./worms.}).

\section{Experiment 1: Replication of \citet{urbach_2010_QuantifiersMoreLess}}

\subsection{Materials}
In this experiment, we use all the stimuli from two experiments carried out by \citet{urbach_2010_QuantifiersMoreLess}. These are made up of 120 sentence frames with 8 different sentence types falling into 4 experimental conditions, for a total of 960 sentences. The 4 conditions had a 2x2 design---each stimulus was either typical (T) or atypical (A), and had either a \textit{most}-type or \textit{few}-type quantifier. An example of the 8 sentence types comprising one sentence frame is shown in (\ref{ex:stimulus_set}).

\begin{subexamples}
\label{ex:stimulus_set}
\item \textit{Most} squirrels gather \textbf{nuts}...  (T, \textit{most})
\item \textit{Most} squirrels gather \textbf{nails}...  (A, \textit{most})
\item \textit{Few} squirrels gather \textbf{nuts}...  (T, \textit{few})
\item \textit{Few} squirrels gather \textbf{nails}...  (A, \textit{few})
\item Squirrels \textit{often} gather \textbf{nuts}...  (T, \textit{most})
\item Squirrels \textit{often} gather \textbf{nails}...  (A, \textit{most})
\item Squirrels \textit{rarely} gather \textbf{nuts}...  (T, \textit{few})
\item Squirrels \textit{rarely} gather \textbf{nails}...  (A, \textit{few})
\end{subexamples} 

The quantifiers used in sentences (a)-(d) differed by sentence frame; see \autoref{sec:appendix_quantifiers} for a full list.

\begin{figure*}[!t]
    \centering
    \includegraphics[width=\textwidth]{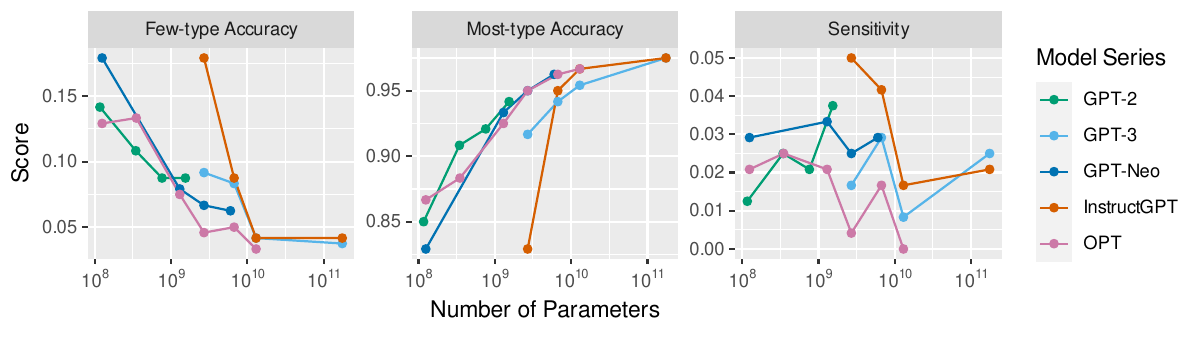}
    \caption{Accuracy and sensitivity of all models.}
    \label{fig:exp1}
\end{figure*}

\subsection{Language Models}

To cover a range of language models with different training data and numbers of parameters, we run our analyses on the GPT-2 \citep{radford_2019_LanguageModelsAre}, GPT-3 \citep{brown_2020_LanguageModelsAre}, GPT-Neo (\citealp{black_2021_GPTNeoLargeScale}; including GPT-J, \citealp{wang_2021_GPTJ6BBillionParameter}), and OPT \citep{zhang_2022_OPTOpenPretrained} language models. We also include an analysis of the first series of InstructGPT models (\texttt{text-davinci-001} etc.), which were finetuned on human-written and highly-rated model-generated responses \citep{openai_2023_ModelIndexResearchers}.

\subsection{Evaluation}
For each stimulus sentence, we calculate the surprisal of the critical word, that is, the word for which the N400 response was measured in the original study. Because humans only encounter the context preceding the critical word when processing the word, and because the language models we analyze are all autoregressive, we only consider the surprisal of the critical word given its preceding context. To do this we truncated the sentence before the critical word, and then used the relevant language model to calculate the probability $p$ of the target word given the preceding context, which was then converted to surprisal $S$ following \autoref{eq:surprisal}.

\setcounter{equation}{0}
\begin{equation}
\label{eq:surprisal}
    S = -\log{p(w_i|w_1...w_{i-1})}
\end{equation}

In previous work of this type \citep[e.g.,][]{ettinger_2020_WhatBERTNot}, only words that were single tokens in the models' vocabularies were used. In this study, all models are autoregressive, so for multi-token words, consecutive sub-word tokens can be predicted, the product of which is a well-defined probability for the whole word. The surprisal of such words, then, is the sum of the surprisals of the sub-word tokens. Calculating surprisal this way allows us to compare the predictions of all the models for all the stimuli in the original experiment.

In order to evaluate how well each model takes into account the quantifier in its predictions, we compared which of the two possible critical words (typical or atypical) had a lower surprisal, i.e., was more strongly predicted by the model. To align with human plausibility judgements, following a \textit{most}-type quantifier, the typical continuation was judged to be correct, and following a \textit{few}-type quantifier, the atypical continuation was judged to be correct. Accuracy was calculated as the fraction of the stimulus pairs for which the model predicted the appropriate critical word---that is, predicted the correct continuation more strongly than the incorrect one. For example, the set of stimuli presented in (\ref{ex:stimulus_set}) is made up of 4 pairs of stimuli, and for a model to achieve 100\% accuracy (4/4), it would need to predict (a) over (b), (d) over (c), (e) over (f), and (h) over (g). This design intrinsically controls for any differences in unconditioned probability among the final words themselves. 

Following \citet{ettinger_2020_WhatBERTNot}, we also analyzed model sensitivity to the quantifiers. In the present study, this corresponds to the question of whether, for a given sentence frame, the model makes a different prediction following a \textit{few}-type quantifier than it does following a \textit{most}-type quantifier. We defined sensitivity as the proportion of stimuli for which the model correctly predicts the critical word following both the \textit{most}-type and the \textit{few}-type quantifier. Thus, the stimuli in each sentence frame provide 2 data points for sensitivity: in (\ref{ex:stimulus_set}), sensitivity is calculated for (a)-(d) and for (e)-(h). For the (a)-(d) stimuli, a model would be considered sensitive to the quantifier if it correctly predicted (a) over (b) \textit{and} (d) over (c). Code and data are available at \url{https://osf.io/vjyw9}.

\subsection{Results}

\label{sec:results}

Each model's accuracy at predicting the critical words following \textit{most}- and \textit{few}-type quantifiers is shown in \autoref{fig:exp1}. All model series show the same general tendencies in accuracy: (1) they perform quite poorly for \textit{few}-type quantifiers but relatively well for \textit{most}-type quantifiers; and (2) as model size increases, word prediction following \textit{most}-type quantifiers improves, but it degrades following \textit{few}-type quantifiers. \autoref{fig:exp1} does show small exceptions to this pattern. From GPT-2 762M to 1542M and from InstructGPT 13B to 175B, while \textit{most}-performance increases, \textit{few}-performance does not decrease. Furthermore, from OPT 125M to 350M, and from OPT 2.7B to 6.7B, there is actually a slight improvement. Nonetheless, these differences are small compared to the overall decreases in performance, and the general trends are still clear---for example, no model performs better on \textit{few}-type quantifiers than a model two or more sizes smaller.

With sensitivity, as shown in \autoref{fig:exp1}, some models improve as they increase in size, and some get worse; however, even the greatest distance between the sensitivity of two models in the same series (InstructGPT 2.7B and 13B) is only 3.4\%. Thus, other than the general fact that sensitivity is low across all models, there does not appear to be any clear pattern, suggesting that sensitivity does not drive the effects seen in accuracy. All accuracy and sensitivity scores can be found in \autoref{sec:appendix_scores}.

\subsection{Discussion}
These results show that contemporary autoregressive transformer models perform poorly on \textit{few}-type quantifiers, and that as these models increase in size, they tend to improve at predicting words following \textit{most}-type quantifiers but get worse at predicting words following \textit{few}-type quantifiers. In fact, we see that models that better predicted the more typical word after a \textit{most}-type quantifier were also worse at predicting the less typical word following a \textit{least}-type quantifier. The fact that models were evaluated on which of the two options they predicted to be more likely, combined with generally poor and largely invariant sensitivity (peaking at 5\%), suggests that the larger models generally made predictions increasingly in accordance with typicality, overwhelming any sensitivity to quantifier type. This aligns with previous work on negation and logical quantifiers in language models \citep{ettinger_2020_WhatBERTNot,kassner_2020_NegatedMisprimedProbes,kalouli_2022_NegationCoordinationQuantifiers}, as well as the N400 results of the original study by \citet{urbach_2010_QuantifiersMoreLess}.

\section{Experiment 2: Sentence-final nouns}

\subsection{Method}
The models and evaluation approach were identical to Experiment 1. The materials were identical to Experiment 1 with the single difference that all nouns were followed by a period, and the surprisal of this period was included when calculating the total surprisal of the critical word (e.g., \textit{nuts.} or \textit{nails.} for the example presented in (\ref{ex:stimulus_set})). Thus, surprisal reflected both the surprisal of the critical word in context and the surprisal of the word being followed by a period, i.e., being the last word in the sentence. For a discussion of modeling the probability of sentence-final words in this way, see \citet{szewczyk_2022_ContextbasedFacilitationSemantic}.

\subsection{Results}
\begin{figure*}[h]
    \centering
    \includegraphics[width=\textwidth]{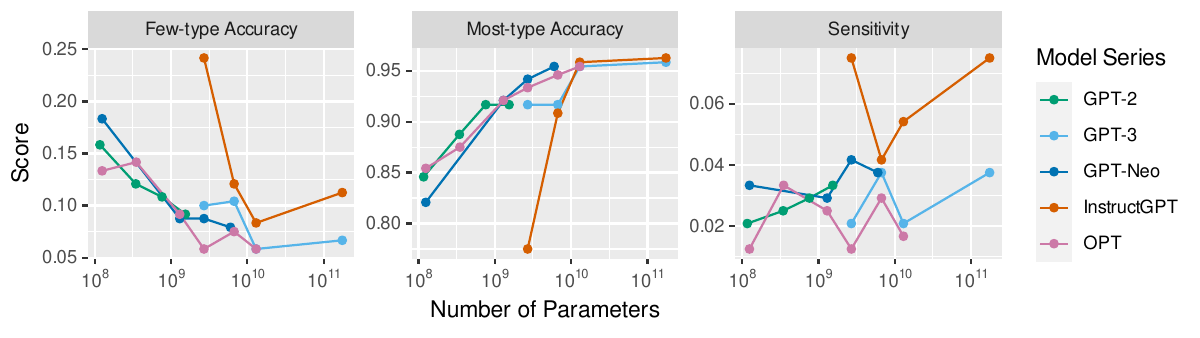}
    \caption{Accuracy and sensitivity of all models on stimuli with added periods (e.g., \textit{Few squirrels gather \textbf{nuts.}}).}
    \label{fig:exp2}
\end{figure*}

Results are shown in \autoref{fig:exp2}. As in Experiment 1, larger models perform worse overall. However, there is a small improvement in the very largest GPT-3 and InstructGPT models relative to the second-largest models of the same type, both in \textit{few}-type accuracy and sensitivity. Performance also increases on these metrics between OPT 2.7B and OPT 6.7B; however, this decreases with OPT 13B. All accuracy and sensitivity scores can be found in \autoref{sec:appendix_scores}.

\subsection{Discusion}
Overall, the results are similar to those of Experiment 1: Larger models of the same type perform worse than smaller models. Whether the small improvement of the largest GPT-3 and InstructGPT models relative to the second-largest models is a fluctuation like that seen for OPT or the beginnings of a U-shaped curve (see \citealp{wei_2022_InverseScalingCan}) is a question for further research.

\section{General Discussion}
In this study, we investigated whether language models show the same insensitivity towards \textit{few}-type and \textit{most}-type quantifiers observed in the predictions made by humans during language comprehension, as indexed by the N400 response. We find that when tested on the same stimuli, they do, predicting the ostensibly implausible \textit{few squirrels gather \textbf{nuts}} to be more likely than \textit{few squirrels gather \textbf{nails}}. Moreover, we find that as language models increase in size, they tend to show this effect to a greater extent, an example of inverse scaling. Based on our analysis of sensitivity and accuracy with \textit{most}-type quantifiers, we hypothesize that these results are due to a low degree of sensitivity to quantifiers and an increase in sensitivity to typicality. In other words, language models appear to be increasingly sensitive to the fact that \textit{squirrels gather \textbf{nuts}} is more plausible than \textit{squirrels gather \textbf{nails}}, but not to the effect on meaning that is caused by a preceding \textit{most} or \textit{few}. 

It is often assumed that as models increase in size and are trained on more data, their performance on natural language tasks generally improves---indeed, evidence supports this \citep{brown_2020_LanguageModelsAre,raffel_2020_ExploringLimitsTransfer,lin_2021_FewshotLearningMultilingual,srivastava_2022_ImitationGameQuantifying,hoffmann_2022_TrainingComputeOptimalLarge,rae_2022_ScalingLanguageModels,zhang_2022_OPTOpenPretrained,chowdhery_2022_PaLMScalingLanguage}. However, the predictions of larger models and those trained on more data also increasingly correlate with human incremental online predictions, in particular those indexed by N400 amplitude \citep{frank_2015_ERPResponseAmount,aurnhammer_2019_ComparingGatedSimple,aurnhammer_2019_EvaluatingInformationtheoreticMeasures,michaelov_2020_HowWellDoes,merkx_2021_HumanSentenceProcessing,michaelov_2021_DifferentKindsCognitive,michaelov_2022_ClozeFarN400}. The two are often aligned---it is easier for humans to process well-formed sentences with plausible semantics \citep{frisch_2005_ResolutionCaseConflicts,nieuwland_2020_DissociableEffectsPrediction}. But in cases such as the present study, the two are not aligned, and we see instead that the predictions of larger models correlate better with human online predictions, even when these are contrary to offline judgements. Thus, the increased performance we see at tasks corresponding to offline human judgements---and note that virtually all manually-annotated tasks are based on offline human judgements---may in fact be a by-product of the models' predictions resembling the online predictions.

Fortunately, the literature boasts a wealth of psycholinguistic studies where metrics of online prediction such as the N400 appear to conflict with offline judgements. Future work could use these to identify phenomena where language models may struggle to make predictions in line with human judgements. Such cases are important to detect as use of LMs becomes more widespread. But by the same token, the present study shows that as language models increase in size, even when augmented by finetuning on desirable responses, they can make predictions that align less and less with explicit human judgements. 

This may be a clear indication of an inherent `outer misalignment' present in language models: while humans might like language models to generate plausible sentences, by their nature they can only generate the most statistically probable ones. Just as there is no guarantee of accuracy or coherence \citep{bender_2021_DangersStochasticParrots}, there is no guarantee of plausibility. While it may be possible to tailor training to avoid specific known issues, this misalignment between probability and plausibility may pose a fundamental challenge with current approaches that aim to use language models as general-purpose natural language systems.

\section*{Limitations}
There are two main limitations to our study. The first is that the stimuli used were limited to those provided by \citeauthor{urbach_2010_QuantifiersMoreLess}'s \citeyearpar{urbach_2010_QuantifiersMoreLess} study. This is because, as stated, we wanted to be able to compare the patterns in the language models' predictions to the patterns in the human N400 response. Thus, we do not look at logical quantifiers like \citet{kalouli_2022_NegationCoordinationQuantifiers}, or any others that have previously been studied \citep[in, e.g.,][]{pezzelle_2018_ThemCanBe,talmor_2020_OLMpicsOnWhatLanguage}.

The other (and perhaps more important) limitation is in the models we were able to use. Crucially, we were not able to access models larger than GPT-3 175B such as PaLM 540B \citep{chowdhery_2022_PaLMScalingLanguage}. This is important because recent work has shown that some inverse scaling patterns become U-shaped (i.e., as language model size increases, performance degrades and then improves again) with such larger models \citep{wei_2022_InverseScalingCan}. 

\section*{Ethics Statement}
Our work complies with the ACL Ethics Policy. Beyond this, we are not aware of any way in which the results of this study may be harmful---in fact, if anything, identifying the limitations of large language models is something that is likely to reduce possible harms by demonstrating cases where their use is not suitable.

From an environmental perspective, we did not train any models; we only used pretrained models for analysis, limiting energy consumption. With the exception of the GPT-3 and InstructGPT models and OPT 13B, all analyses were run on an NVIDIA RTX A6000 GPU, taking a total of 43 minutes. OPT 13B was too large to run on this GPU, and thus was run on an Intel Dual Xeon E7-4870 CPU for a total of 22 hours and 39 minutes. Finally, the GPT-3 and the InstructGPT models were run using the OpenAI API, and thus we do not have access to information about the GPUs used.

\section*{Acknowledgements}
We would like to thank the anonymous reviewers for their helpful comments. We would also like to acknowledge the other members of the Language and Cognition Lab at UCSD for their valuable discussion, as well as Roger Levy and attendees of the MIT Computational Psycholinguistics Laboratory meeting. Finally, we would like to thank the San Diego Social Sciences Computing Facility Team for the use of the Social Sciences Research and Development Environment (SSRDE) cluster. The RTX A6000 used for this research was donated by the NVIDIA Corporation.

\bibliography{library}
\bibliographystyle{acl_natbib}

\clearpage
\appendix

\section{Scores}
\label{sec:appendix_scores}
The performance of each model is presented in \autoref{tab:all_data}.

\begin{table}[!h]
\renewcommand{\arraystretch}{1.2}
\centering
\begin{small}
\begin{tabular}{lrrlrlllll}
\hline
                                                & \multicolumn{4}{c}{\textbf{Critical word}}                                                                                                         &                      & \multicolumn{4}{c}{\textbf{Critical word + period}}                                                                                                \\ \cline{2-5} \cline{7-10} 
\textbf{}                                       & \multicolumn{2}{c}{\textbf{Accuracy}}                                                  & \multicolumn{1}{c}{} & \multicolumn{1}{c}{\textbf{Sens.}} & \multicolumn{1}{c}{} & \multicolumn{2}{c}{\textbf{Accuracy}}                                                  & \multicolumn{1}{c}{} & \multicolumn{1}{c}{\textbf{Sens.}} \\ \cline{2-3} \cline{5-5} \cline{7-8} \cline{10-10} 
\textbf{Model}                                  & \multicolumn{1}{c}{\textit{\textbf{most}}} & \multicolumn{1}{c}{\textit{\textbf{few}}} & \multicolumn{1}{c}{} & \multicolumn{1}{c}{\textbf{}}      & \multicolumn{1}{c}{} & \multicolumn{1}{c}{\textit{\textbf{most}}} & \multicolumn{1}{c}{\textit{\textbf{few}}} & \multicolumn{1}{c}{} & \multicolumn{1}{c}{\textbf{}}      \\ \hline
GPT-2 117M (\texttt{gpt2})                      & 0.850                                      & 0.142                                     &                      & 0.013                              &                      & 0.846                                      & 0.158                                     &                      & 0.021                              \\
GPT-2 345M (\texttt{gpt2-medium})               & 0.908                                      & 0.108                                     &                      & 0.025                              &                      & 0.887                                      & 0.121                                     &                      & 0.025                              \\
GPT-2 762M (\texttt{gpt2-large})                & 0.921                                      & 0.088                                     &                      & 0.021                              &                      & 0.917                                      & 0.108                                     &                      & 0.029                              \\
GPT-2 1542M (\texttt{gpt2-xl})                  & 0.942                                      & 0.088                                     &                      & 0.038                              &                      & 0.917                                      & 0.092                                     &                      & 0.033                              \\ \hline
GPT-3 2.7B (\texttt{ada})                       & 0.917                                      & 0.092                                     &                      & 0.017                              &                      & 0.917                                      & 0.1                                       &                      & 0.021                              \\
GPT-3 6.7B (\texttt{babbage})                   & 0.942                                      & 0.083                                     &                      & 0.029                              &                      & 0.917                                      & 0.104                                     &                      & 0.038                              \\
GPT-3 13B (\texttt{curie})                      & 0.954                                      & 0.042                                     &                      & 0.008                              &                      & 0.954                                      & 0.058                                     &                      & 0.021                              \\
GPT-3 175B (\texttt{davinci})                   & 0.975                                      & 0.038                                     &                      & 0.025                              &                      & 0.958                                      & 0.067                                     &                      & 0.038                              \\ \hline
InstructGPT 2.7B (\texttt{text-ada-001})        & 0.829                                      & 0.179                                     &                      & 0.050                              &                      & 0.775                                      & 0.242                                     &                      & 0.075                              \\
InstructGPT 6.7B (\texttt{text-babbage-001})    & 0.950                                      & 0.088                                     &                      & 0.042                              &                      & 0.908                                      & 0.121                                     &                      & 0.042                              \\
InstructGPT 13B (\texttt{text-curie-001})       & 0.967                                      & 0.042                                     &                      & 0.017                              &                      & 0.958                                      & 0.083                                     &                      & 0.054                              \\
InstructGPT 175B (\texttt{text-davinci-001})    & 0.975                                      & 0.042                                     &                      & 0.021                              &                      & 0.963                                      & 0.112                                     &                      & 0.075                              \\ \hline
GPT-Neo 125M (\texttt{EleutherAI/gpt-neo-125m}) & 0.829                                      & 0.179                                     &                      & 0.029                              &                      & 0.821                                      & 0.183                                     &                      & 0.033                              \\
GPT-Neo 1.3B (\texttt{EleutherAI/gpt-neo-1.3B}) & 0.933                                      & 0.079                                     &                      & 0.033                              &                      & 0.921                                      & 0.088                                     &                      & 0.029                              \\
GPT-Neo 2.7B (\texttt{EleutherAI/gpt-neo-2.7B}) & 0.950                                      & 0.067                                     &                      & 0.025                              &                      & 0.942                                      & 0.088                                     &                      & 0.042                              \\
GPT-J 6B (\texttt{EleutherAI/gpt-j-6b})         & 0.963                                      & 0.062                                     &                      & 0.029                              &                      & 0.954                                      & 0.079                                     &                      & 0.038                              \\ \hline
OPT 125M (\texttt{facebook/opt-125m})           & 0.867                                      & 0.129                                     &                      & 0.021                              &                      & 0.854                                      & 0.133                                     &                      & 0.013                              \\
OPT 350M (\texttt{facebook/opt-350m})           & 0.883                                      & 0.133                                     &                      & 0.025                              &                      & 0.875                                      & 0.142                                     &                      & 0.033                              \\
OPT 1.3B (\texttt{facebook/opt-1.3b})           & 0.925                                      & 0.075                                     &                      & 0.021                              &                      & 0.921                                      & 0.092                                     &                      & 0.025                              \\
OPT 2.7B (\texttt{facebook/opt-2.7b})           & 0.950                                      & 0.046                                     &                      & 0.004                              &                      & 0.933                                      & 0.058                                     &                      & 0.013                              \\
OPT 6.7B (\texttt{facebook/opt-6.7b})           & 0.963                                      & 0.050                                     &                      & 0.017                              &                      & 0.946                                      & 0.075                                     &                      & 0.029                              \\
OPT 13B (\texttt{facebook/opt-13b})             & 0.967                                      & 0.033                                     &                      & 0                                  &                      & 0.954                                      & 0.058                                     &                      & 0.017                              \\ \hline
\end{tabular}
\end{small}
\caption{Accuracy and sensitivity scores for all models.}
\label{tab:all_data}
\end{table}

\section{Quantifiers}
\label{sec:appendix_quantifiers}
\autoref{tab:quantifiers} lists all quantifiers used and the proportion of sentences in WikiText-103 that contain them.

\begin{table*}[!b]
\renewcommand{\arraystretch}{1.2}
\centering
\begin{small}
\begin{tabular}{llllll}
\hline
               & \multicolumn{2}{c}{\textbf{\textit{Most}-type}}               & \multicolumn{1}{c}{} & \multicolumn{2}{c}{\textbf{\textit{Few}-type}}                \\ \cline{2-3} \cline{5-6} 
               & \textbf{Quantifier} & \textbf{Frequency (sentences)} &                      & \textbf{Quantifier} & \textbf{Frequency (sentences)} \\ \hline
               & most                & 0.025177                       &                      & few                 & 0.005870                       \\
               & almost all          & 0.000305                       &                      & almost no           & 0.000098                       \\
               & practically all     & 0.000009                       &                      & practically no      & 0.000008                       \\
               & a large number of   & 0.000300                       &                      & a small number of   & 0.000131                       \\
               & nearly all          & 0.000170                       &                      & rather few          & 0.000001                       \\
               & lots of             & 0.000153                       &                      & hardly any          & 0.000017                       \\
               & a lot of            & 0.000745                       &                      & a very few          & 0.000010                       \\
               & many                & 0.015874                       &                      & few                 & 0.005870                       \\
               & often               & 0.005766                       &                      & rarely              & 0.000610                       \\ \hline
\textbf{Total} &                     & 0.046809                       &                      &                     & 0.006717                       \\ \hline
\end{tabular}
\end{small}
\caption{In each sentence frame, \textit{most} and \textit{few}-type quantifiers were matched based on their meanings as length in number of words \citep{urbach_2010_QuantifiersMoreLess}. Matched quantifiers are shown beside each other. As can be seen, \textit{few} is matched to both \textit{most} and \textit{many}. The frequency of each quantifier is given in terms of the proportion of sentences in WikiText-103 \citep{merity_2017_PointerSentinelMixture} that contain it. The total frequencies are the number of sentences in WikiText-103 that contain at least one of either the \textit{few}-type or \textit{most}-type quantifiers; not the sum of the individual quantifier frequencies.}
\label{tab:quantifiers}
\end{table*}

\end{document}